%% file: blitz.tex
\documentclass[nosumlimits]{article} 

\usepackage{hyperref}
\usepackage{url}
\usepackage[margin=1.5in]{geometry}
\usepackage[xindy]{glossaries}
\include{gloss}
\usepackage{amsmath}
\usepackage{amsfonts}
\usepackage{tikz}
\usepackage{pgfplots}
\pgfplotsset{compat=1.5}
\usepackage{amssymb}
\usepackage{xfrac}
\usepackage{listings}
\usepackage{graphicx}
\usepackage{bm}
\usepackage{cleveref}
\usepackage{booktabs}
\usepackage{subcaption} 
\usepgfplotslibrary{external}%
\tikzset{external/system call={xelatex \tikzexternalcheckshellescape
   -halt-on-error -interaction=batchmode -jobname "image" "texsource"}}

\newcommand{\bs}[1]{\mathbf{#1} }
\crefname{equation}{equation}{equations}
\crefname{figure}{figure}{figures}
\crefname{table}{table}{tables}

\usepackage{mathtools}
\newcommand{\defeq}{\coloneqq}

\title{Blitzkriging: Kronecker-structured Stochastic Gaussian Processes}

\author{
Thomas Nickson \ \ \ Tom Gunter \ \ \ Chris Lloyd  \\  Michael A Osborne \ \ \ Stephen Roberts \\
Department of Engineering Science\\
University of Oxford\\
Oxford, UK \\
\texttt{\{tron,tbgunter,clloyd,mosb,sjrob\}@robots.ox.ac.uk} \\
}

\begin{document}

\maketitle

\begin{abstract}
We present Blitzkriging, a new approach to fast inference for Gaussian processes, applicable to regression, optimisation and classification. 
State-of-the-art (stochastic) inference for Gaussian processes on very large datasets scales cubically in the number of `inducing inputs', variables introduced to factorise the model. 
Blitzkriging shares state-of-the-art scaling with data, but reduces the scaling in the number of inducing points to approximately linear. 
Further, in contrast to other methods, Blitzkriging: does not force the data to conform to any particular structure (including grid-like); reduces reliance on error-prone optimisation of inducing point locations; and is able to learn rich (covariance) structure from the data.
We demonstrate the benefits of our approach on real data in regression, time-series prediction and signal-interpolation experiments. 
\end{abstract}
\glsresetall

\section{Introduction and motivation}

The \gls{gp} is a ubiquitous method for functional inference in Bayesian  machine learning. It is typically used: for regression (where it is also known as \textit{kriging}) and classification \cite{Rasmussen:2005:GPM:1162254}; in non-linear dimensionality reduction as the \gls{gplvm} \cite{Lawrence:2005:PNP:1046920.1194904}; as a prior over the rate function in Cox processes \cite{point_process} and broadly within Probabilistic Numerics (see \url{probabilistic-numerics.org}).

While these methods have shown class-leading performance on small datasets, the trend in industry and academia has been towards larger, noisier and more redundant data. The success of deep networks and ensemble methods such as the random forest can be credited to their ability to generalise and learn when very large volumes of data are available. In this paper, we present an approximation to the \gls{gp} that retains strong performance on small datasets, whilst benignly scaling to large datasets.

\section{The \glsentrylong{gp}}
\label{gp}

For a domain $\mathcal{X} = \mathbb{R}^D$, the \gls{gp} is a prior over a stochastic scalar function defined as  $f(x) : \mathcal{X}  \rightarrow \mathbb{R}$.  This means that the prior for any finite set of function values is multivariate Gaussian, defined by a positive definite covariance function $C(\cdot,\cdot) : \mathcal{X} \times \mathcal{X} \rightarrow \mathbb{R}$ and a mean function $m(\cdot) : \mathcal{X} \rightarrow \mathbb{R}$. The mean and covariance function are parameterised by a set of hyperparameters. For the purposes of this paper, we will assume that observations are possibly i.i.d. Gaussian noise corrupted values $\bs{y}$ of $f$ at $\bs{x}$. A good introduction to the \gls{gp} can be found in \cite{Rasmussen:2005:GPM:1162254}.

Given $N$ function evaluations at $\{\mathbf{\bs{x}_i}\}_{i=1}^N$, inference with \Glspl{gp} is generally computationally bottlenecked by linear algebra operations on an $N \times N$ dense covariance (function) matrix. Inverting this matrix has complexity $\mathcal{O}\left( N^3 \right)$, and restricts the \gls{gp} to applications where $N \lesssim  10,000$ on modern hardware. Due to this restriction, there has been much work on finding approximations to the \gls{gp} that preserve the rich structure but have more tractable computational requirements. 

\section{Tractable \glsentrylong{gp}es}
\paragraph{Woodbury methods} The majority of \gls{gp} approximations in the literature rely on the Woodbury matrix identity, 
\begin{equation}\textstyle	\left(\bs{A} + \bs{U}\bs{C}\bs{V}\right)^{-1}
	=
	\bs{A}^{-1} -
	\bs{A}^{-1}\bs{U} \left(
	\bs{C}^{-1}
	+
	\bs{V}\bs{A}^{-1}\bs{U}
	\right)^{-1}
	\bs{V}\bs{A}^{-1}.
\end{equation}

to reduce the complexity of the calculations. One popular family of approximations exploiting this identity is reviewed by Quinonero-Candela \cite{Quinonero-Candela:2005:UVS:1046920.1194909}. These methods make use of `inducing inputs', a reduced set of pseudo-observations where the function values at these points are analytically marginalised. Taking $M$ inducing inputs, over which there is an  $M \times M$ covariance matrix $\bs{K}_{mm}$, the full Gram matrix is approximated as 
$\bs{K}_{nn} \approx \bs{K}_{nm} \bs{K}_{mm}^{-1} \bs{K}_{mn}$, where $\bs{K}_{nm}$ is the $N \times M$ covariance matrix between the data and the inducing points.

This approximation allows the application of the Woodbury identity. With i.i.d. Gaussian noise of variance $\beta^2$:
\begin{equation}\textstyle
\bs{K}_{nn} + \beta^2 \bs{I} \approx  \beta^2 \bs{I} + \bs{K}_{nm} \bs{K}_{mm}^{-1} \bs{K}_{mn} \defeq \bs{A} + \bs{U}\bs{C}\bs{V}.
\end{equation}

An alternative approach to reducing the rank of the matrix inversion is to find an orthogonal basis function representation of the kernel, such as 
the sparse spectrum \gls{gp} by Lazaro-Gredilla et al. \cite{sparsespec}.
 Once the low rank representation of the kernel has been achieved, inference proceeds as above using the Woodbury identity.

\paragraph{Product of Experts} Products of experts train many \glspl{gp} models on subsets of the data of size $S$, giving a cost that is proportional to $\mathcal{O}(S^3)$. These subsets reduce the expressivity of the model, however Deisenroth and Ng \cite{dgp} have had success scaling these models to large data sets.

\paragraph{Kronecker methods} Another framework for fast multi-dimensional \gls{gp} regression exploits the properties of product kernels (taken as a product of kernels, one over each input separately) on regular grids, which can be represented by the Kronecker product. A full explanation of this approach can be found in Saat\c{c}i's thesis, \cite{Saatci11}. This  structure allows inference to be performed in linear time with data, as long as the data lies on a grid (examples of gridded data can be found in time series, videos and images). Notably, the \gls{gp} is not approximated and inference is exact. A nearly full grid can be enhanced with pseudo-observations with near-infinite noise with little effect on the accuracy, as in  \textit{GPatt} by Wilson et al. \cite{wilson2013gpatt}.

The dense grid of this formulation allows the use of rich kernels, such as the \gls{sm} kernel introduced by Wilson et al. \cite{spectralkernel}. The spectrum of this kernel is parameterised as a sum of Gaussians, providing an everywhere positive function. 
By Bochner's theorem \cite{Rasmussen:2005:GPM:1162254}, this is also a valid covariance, and has the following form:
\begin{equation}
\textstyle
k_{sm}(\bm{x}, \bm{x'}) = \prod_{d=1}^D \sum_{q=1}^Q w_q^2 \exp\left(- 2 \pi ^ 2 | x _d- x'_d |^2 \sigma_q ^2 \right)\cos\left(2 \pi | x_d - x'_d | \mu_q\right).
  \label{eq:spectral}
\end{equation}
The locations, widths and magnitudes of these functions are the hyper-parameters of the kernel, and can be optimised or marginalised. 
This allows complex and long-range kernels to be learned based on the data. 

\paragraph{Kernel Interpolation for Scalable Structured Gaussian Processes}

The KISS-GP (Kernel Interpolation for Scalable Structured Gaussian Processes)\cite{DBLP:journals/corr/WilsonN15} is an inducing point method which approximates the cross-covariance matrix $\mathbf{K}_{xu}$ between the observed data and the inducing points with a weighted sum of entries from the inducing point covariance $\mathbf{K}_{uu}$. The cross-covariance is approximated as $\mathbf{K}_{xu} \approx \mathbf{W}\mathbf{K}_{uu}$, where $\mathbf{W}$ is a sparse matrix containing 2 or 4 weights. The inducing points can be placed in arbitrary positions, which allows the exploitation of Kronecker structure in the inducing space while avoiding the restriction on data location mentioned above. This method is similar to the method presented in this paper, in that both restrict the structure of the inducing space for computational gains.


\paragraph{Variational methods}
The inducing point and sparse spectrum methods above are prone to overfitting. Titsias \cite{Titsias09variationallearning} and Gal and Turner \cite{variational_sparsespec} address this by representing the uncertainty in the inducing and spectral points, respectively, with a variational distribution.

\paragraph{Stochastic variational methods} The basis function and inducing points methods above, including the variational \gls{fitc} approach of Titsias, all scale as $\mathcal{O}(N M^2)$ for $N$ data and $M$ inducing points. Additionally, the likelihood can not be split up over the data, precluding the use of stochastic gradient methods that allow learning on large datasets, through cheap but noisy observations of the gradient of the objective function on minibatches (small subsets) of the full data.

Both of these issues have been solved in Hensman et al. \cite{bigdatagp} with their \gls{svgp}
and a generalisation to all inducing methods by Hoang et al. \cite{generalsparse}. We concentrate on the \gls{svgp} here, for the ease of testing against the popular GPy library \cite{gpy2014}.
The \gls{svgp} does not marginalise the function values $\bs{u}$ at the inducing point locations and represents them with a further variational distribution. This approach yields a lower bound on the lower bound on the log-likelihood $\log p(\bm{y} \mid \bs{X})$ derived by Titsias \cite{Titsias09variationallearning}:

\begin{equation}
  \log p(\bm{y} \mid \bs{X})
  \geq
  \mathcal{L}_3 
  \defeq
  \sum\limits_{i=1}^N 
  \biggl(
  \log\mathcal{N}(y_i \mid \mathbf{k}_i^T \mathbf{K}_{mm}^{-1} \mathbf{m}, \beta^{-1}) 
  - 
  \frac{1}{2}\beta \tilde{k}_{i,i} 
  -
  \frac{1}{2}\text{tr}\left(\mathbf{S}\bm{\Lambda}_i\right)
  \biggr)
  -
  D_\text{KL}\bigl(q(\mathbf{u}) \| p(\mathbf{u})\bigr)
  \label{eq:l3}
\end{equation}

where $\bm{\Lambda}_i = \beta\bs{K}_{mm}^{-1} \bs{k}_i \bs{k}_i^T \bs{K}_{mm}^{-1}$, $\bs{k}_i$ is the $i^{\text{th}}$ column of $\bs{K}_{mn}$, $\tilde{k}_{i,i}$ is the $i^{\text{th}}$ diagonal of $\bs{K}_{nn} - \bs{K}_{nm}\bs{K}_{mm}^{-1}\bs{K}_{mn}$, $q(\mathbf{u})$ is the variational distribution $\mathcal{N}\left(\bs{u} | \bs{m}, \bs{S} \right)$, $p(\bs{u})$ is the \gls{gp} prior and $D_\text{KL}\bigl(q(\mathbf{u}) \| p(\mathbf{u})\bigr) $ is the \gls{kl} divergence between $q(\mathbf{u})$ and $p(\mathbf{u})$.

By setting the variational mean, $\bs{m}$, and covariance, $\bs{S}$, as
\begin{equation}
  \bs{m} = \beta \bm{\Lambda}^{-1}\bs{K}_{mm}^{-1}\bs{K}_{mn}\bs{y} 
  \qquad \text{and} \qquad
  \bs{S} = \bm{\Lambda}^{-1},
  \label{eq:optS}
\end{equation}
where $\bm{\Lambda} = \beta \bs{K}_{mm}^{-1} \bs{K}_{mn}\bs{K}_{nm}\bs{K}_{mm}^{-1} + \bs{K}_{mm}^{-1}$, the original bound by Titsias \cite{Titsias09variationallearning} is recovered. Further, placing the inducing points at the training data locations recovers the full \gls{gp}, although doing so erases any computational gains.

The unwrapping of the likelihood in \cref{eq:l3} allows it to decompose over the data, a necessary property for distributed computation and stochastic inference. The complexity is $\mathcal{O}(BM^3)$ for a mini-batch size $B$. The variational distribution of the inducing point values is represented by a mean vector $\bs{m}$ and variational covariance $\bs{S}$, parameterised using its lower triangular Cholesky factor $\bs{L}$. 

\section{Kronecker structured stochastic variational Gaussian processes} 


We present a stochastic, variational, Kronecker-structured \gls{gp} that we term {\textit{Blitzkriging}}.
Kronecker structure can only be exploited in the full \gls{gp} when the data lies on a grid and is modelled by a product kernel. The \gls{svgp} bound in \cref{eq:l3} can exploit Kronecker structure for real computational gains if: the inducing points are placed on a grid; the variational covariance is taken as $\bs{S} = \bigotimes_{d=1}^D \bs{S}_{m_d m_d}$, and; the kernel is a product kernel, $\bs{K}_{mm} = \bigotimes_{d=1}^D \bs{K}_{m_d m_d}$.
Blitzkriging is precisely the \gls{svgp} modified in this way.  

This modification to the approximate bound means that it cannot be optimal. The optimal setting for the variational covariance $\bs{S}$ is given in \cref{eq:optS} as $(\beta \bs{K}_{mm}^{-1} \bs{K}_{mn}\bs{K}_{nm}\bs{K}_{mm}^{-1} + \bs{K}_{mm}^{-1})^{-1}$, which \emph{cannot} have Kronecker structure. The variational machinery of {Blitzkriging} finds the closest approximation to this optimal setting for $\bs{S}$ that factorises as a Kronecker matrix. We show empirically throughout this paper that this approximation does not impact performance on real and generated data.
%
%

This factorisation allows us to simplify expensive operations such as matrix inversion, multiplication and finding the Cholesky factors, because $ (\bs{A}_0 \otimes \bs{A}_1)(\bs{B}_0 \otimes \bs{B}_1) = \bs{A}_0 \bs{B}_0 \otimes \bs{A}_1 \bs{B}_1$. With careful book-keeping, we are able to ensure that we \emph{never} have to perform any operations on the full $M \times M$ inducing point matrix. This allows the use of many more inducing points than possible under the non-Kronecker \gls{svgp}, with improved numerical stability and computational complexity. As noted by Wilson in {GPatt} \cite{wilson2013gpatt}, the standard inducing point approximations do not allow sufficient inducing points to constrain rich kernels such as the \gls{sm} kernel mentioned above. The light-weight but dense inducing grids available in {Blitzkriging} enable us to deploy the \gls{sm} kernel on arbitrary data at scale. 
Crucially, note that we do not require the data to conform to any particular structure.

\subsection{Scaling}
\label{scaling}

\paragraph{Notation} We denote a full matrix by $\bs{M}$, and its per dimension components as $\bs{M}_d$ for the $d^{th}$ dimension. We denote the size $N$ of an $N \times N$ square matrix as $\mathrm{n}(\bs{M})$. The kernel matrix and the variational covariance are always stored as a list of per-dimension matrices. The cross-covariance matrix $\bs{K}_{nm}$ is represented as a partitioned matrix where the rows are the Kronecker product of vectors (we will refer to this by its proper name of a Khatri-Rao matrix), again stored in a list. It in important to note that these matrices are never explicitly evaluated.

%
%

%

\paragraph{Kronecker matrix, full vector product}

In many cases it is necessary to calculate $(\bigotimes_{d=1}^D \bs M_d) \bs v,
$ where $\bs v$ is a full vector of size $\prod_{d=1}^D\mathrm{n}\left(\bs M_d\right)$. Due to the tiled structure of the Kronecker matrix, it is possible to avoid the usual $\mathcal{O}\left(N^2\right)$ operations by noting that for each matrix only $\mathcal{O}(\mathrm{n}\left(\bs M_d\right) \times N)$ unique operations must be performed. In their 1998 paper, Fernandes, Plateau and Stewart \cite{kronmul} provide an algorithm to perform this calculation. This result was noted by Saat\c{c}i \cite{Saatci11}, who provides a alternative way to perform this operation using optimised \gls{blas} operations and matrix transposes. In total, the complexity of this operation 
with an equal  number of inducing points per dimension is 
$
	\mathcal{O}(
	\mathrm{n}\left(\bs M\right) ^{D + 1}
	).$



\paragraph{Khatri-Rao matrix, full vector product} 

While there is no structure here to allow repeated operations to be avoided, there is still a saving in memory usage compared to evaluating the full Khatri-Rao matrix. The overall complexity is $\mathcal{O}(B \mathrm{n}\left(\bs M\right)^D)$, where $B$ is the size of the minibatch we have chosen to evaluate. We can again apply the tensor algebra method from Saat\c{c}i \cite{Saatci11}, to calculate each of the $1 \times M$ Kronecker matrix-full vector products using fast \gls{blas} kernels. In the code used for this paper we found our multi-threaded C code to be faster for the majority of inducing point configurations due to it avoiding a tight loop in Python, however there is a clear path to improve speed.

\paragraph{Diagonal of Khatri-Rao matrix, Kronecker matrix quadratic}

The diagonal of the product of a Khatri-Rao matrix $\bs{M}$ and a matrix with Kronecker structure $\bs{K}^{-1}$, $\mathrm{diag}\left(\bs{M}^\top \bs{K}^{-1} \bs{M}\right)$ can be evaluated by first noting that the product of a Khatri-Rao matrix with a Kronecker matrix is a Khatri-Rao matrix. For each element of the diagonal the sum of the product of the Kronecker structured vectors (\cref{eq:nodiag}) must be evaluated, taking $\mathcal{O}\left(N D \mathrm{n}\left(\bs M\right)\right)$ time. 
\begin{equation}
\textstyle
	\bs A_{ii} 
	= \sum \left( \bs v_{i0} \otimes \bs v_{i1} \right) \odot \left( \bs w_{i0} \otimes \bs w_{i1} \right)
	=  \sum \left( \bs v_{i0} \odot \bs w_{i0} \right) \sum \left( \bs v_{i1} \odot \bs w_{i1} \right),
	\label{eq:nodiag}
\end{equation}

The most expensive operation is the Khatri-Rao matrix, full vector product. This is used in the predictive equation and to calculate the accuracy of the data fit in \cref{eq:l3}.

\subsection{Variational Posterior}
\label{variational_post}
\begin{figure}
\centering
\begin{subfigure}{0.45\textwidth}
  \centering
  \resizebox{0.9\textwidth}{!}{
  \includegraphics{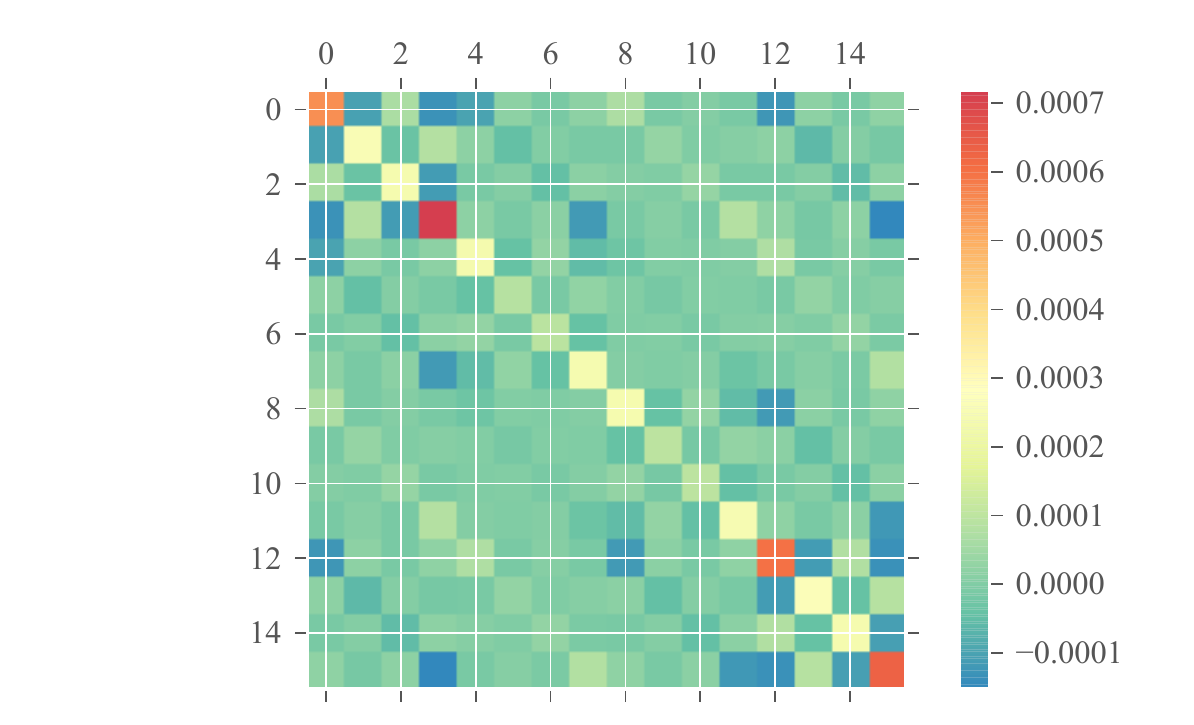}
  }
  \caption{Variational covariance of the \gls{svgp}.}
  \label{fig:cov_svgp_diag}
\end{subfigure}%
\hspace{2em}
\centering
\begin{subfigure}{0.45\textwidth}
  \centering
  \resizebox{0.9\textwidth}{!}{
  \includegraphics{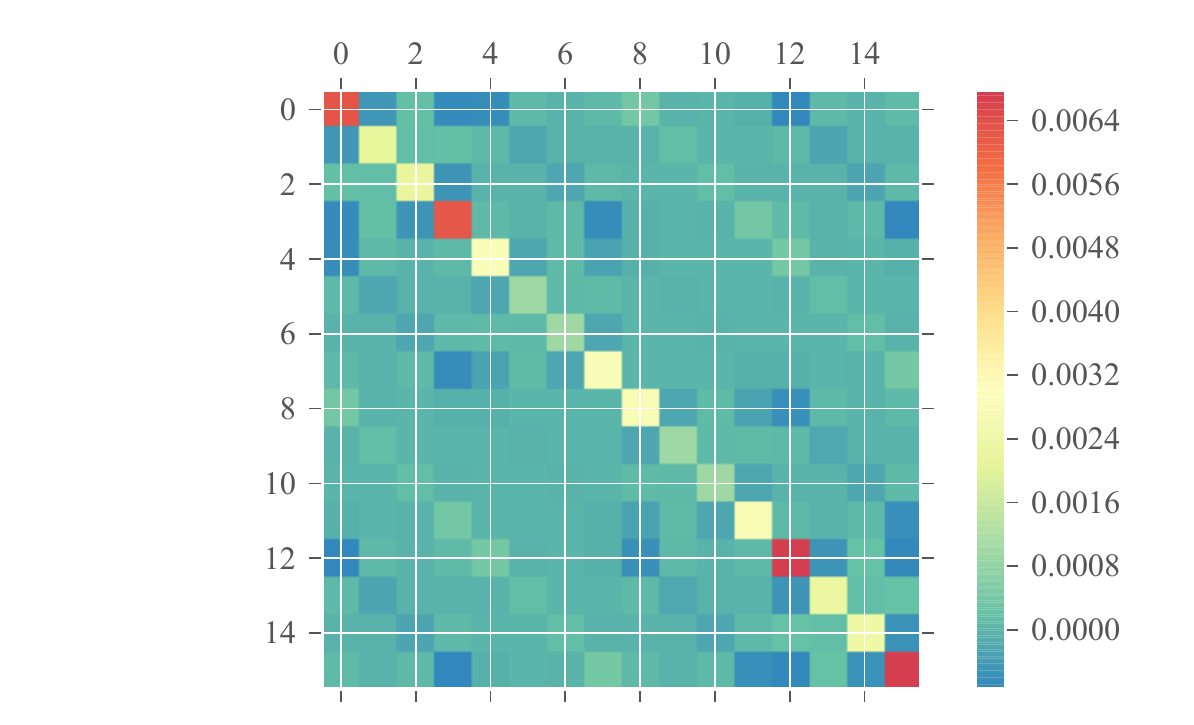}
  }
  \caption{Variational covariance of {Blitzkrige}.}
  \label{fig:cov_blitz_diag}
\end{subfigure}%

\begin{subfigure}{0.45\textwidth}
  \centering
  \resizebox{0.9\textwidth}{!}{
  \includegraphics{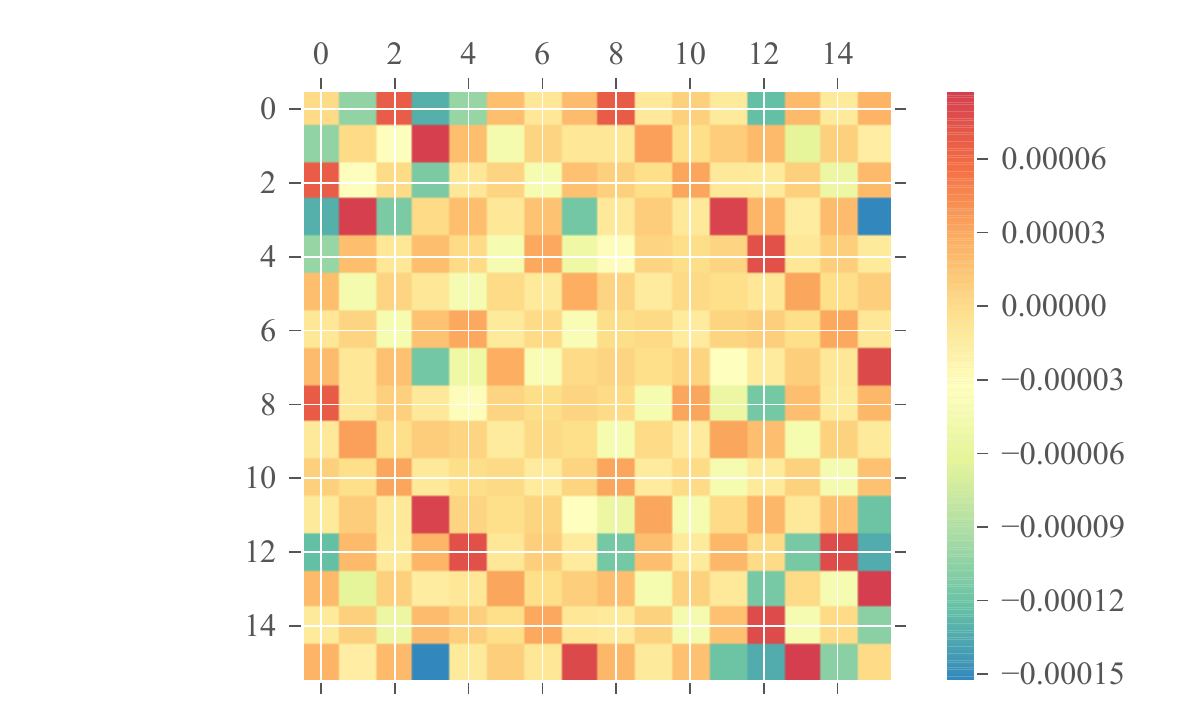}
  }
  \caption{Variational covariance of the \gls{svgp} with diagonal elements set to 0.}
  \label{fig:cov_svgp_nodiag}
\end{subfigure}
\hspace{2em}
\begin{subfigure}{0.45\textwidth}
  \centering
  \resizebox{0.9\textwidth}{!}{
  \includegraphics{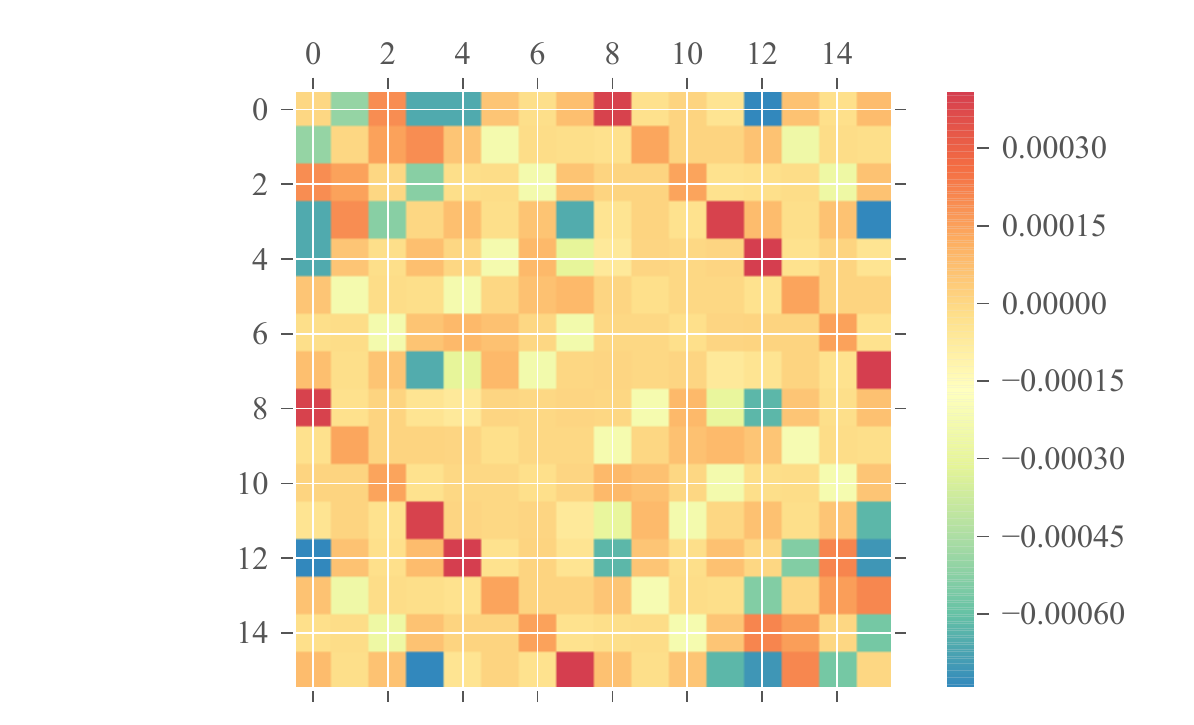}
  }
  \caption{Variational covariance of {Blitzkrige} with diagonal elements set to 0.}
  \label{fig:cov_blitz_nodiag}
\end{subfigure}\par\medskip

\label{fig:both_cov}
\caption{\Gls{svgp} and {Blitzkrige} variational covariance on a two dimensional function with a $4 \times 4$ grid. In the first row, we show that Blitzkrige learns a stronger diagonal variance, implying it is allowing for a less accurate model fit by inflating the uncertainty. The second row shows that the structure of the off-diagonal covariance is similar between both models, though {Blitzkrige} once again infers more extreme values for the elements.}

\end{figure}

{Blitzkriging} forces the variational covariance of the \gls{svgp} to factorise over the input dimensions. The original \gls{svgp}  does not make this factorisation assumption, allowing a more complex covariance structure to be learned.
\Cref{fig:both_cov} shows the effects of this approximation to the covariance. To generate these figures, we learned the \gls{svgp} and \textit{\textit{\textit{Blitzkrige}}} on a complex function evaluated on a $4 \times 4$ grid. In \cref{fig:cov_svgp_diag,fig:cov_blitz_diag}, the ratios of the elements of the diagonals are similar between the \gls{svgp} and \textit{Blitzkriging} (inducing points 0, 3, 12 and 15 have relatively higher variance, while 5, 6, 8 and 9 have lower), however, \textit{Blitzkriging} generally learns stronger variances. Removing the diagonals (\cref{fig:cov_svgp_nodiag,fig:cov_blitz_nodiag}), the same relationship can be seen: a similar structure, with stronger covariance in \textit{Blitzkriging}.

\section{Illustrative examples}

\paragraph{In-model data}
\label{toy}
We tested {Blitzkriging} against the non-Kronecker \gls{svgp} on data drawn from a \gls{gp} model with an \gls{eq} kernel, with an output scale of 1 and and lengthscale $\sqrt{20}$.
We generated a vector of 5,500 points, uniformly distributed in $x \in [-2, 2], y \in [-2, 2]$. 5,000 of these points were used for training, and 500 for testing. 

We tested {Blitzkriging}, the full \gls{gp} model and the standard \gls{svgp} with the \gls{eq} kernel on this data. {Blitzkriging} performed best with the inducing points locked to a grid, while the \gls{svgp} performed best with them optimised jointly with the other parameters. We tested both models with 1 to 1,225 inducing points, {Blitzkriging} increasing as $n = 1, 4, 9,\dots$  and the \gls{svgp} as $n = 1, 11, 21,\dots$. We defined the inducing points on an uniform two dimensional grid for {Blitzkriging}, and drew them from the same distribution as the data for the \gls{svgp} and optimised jointly with the other parameters.

The \gls{svgp} suffered from numerical stability issues when using more than 120 inducing points. {Blitzkriging} did not suffer from this, being able to use up to the maximum number of inducing points and nearly matching the \gls{gp}'s performance in half the run time. The peak performance of {Blitzkrige} was with a $27 \times 27$ inducing point grid, returning an average predictive log-likelihood of 97.2 over 5 runs, compared to the full \gls{gp} which scored an average of 122. Even on this relatively small problem, the computational simplicity of our model compared to the full \gls{gp} results in {Blitzkriging}'s average run time being just over half that of the \gls{gp}, 216 seconds compared to 389.


\paragraph{Pattern discovery on free-form data}
\label{pattern}
In this section we provide an illustrative example of our model finding local and long range periodic structure from non gridded data. The \gls{sm} kernel offers a way to learn the covariance structure from the data, as opposed to expert selection of kernels. Wilson et al. \cite{spectralkernel} show that the \gls{sm} kernel is able to find long range correlations in the data, but is limited to data lying on a (mostly) complete grid.  

Our model removes the requirement for the data to be gridded, allowing pattern discovery on non-gridded data by inferring a gridded representation in the latent inducing space. In {GPatt} \cite{wilson2013gpatt}, Wilson et al. argue inducing point approximations are unable to capture the necessary information.  This is true for conventional inducing point methods, however, {Blitzkriging} is able to use sufficient points to capture complex, long-range, correlations.

\begin{figure}[h]
\centering
\begin{subfigure}{0.48\textwidth}
  \centering
  \resizebox{0.6\textwidth}{!}{
  \includegraphics{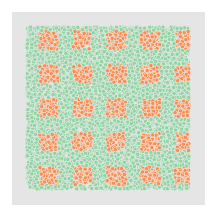}
  }
  \caption{A sharp edged and periodic signal, evaluated at 10,000 draws from a uniform random distribution. 
  }
  \label{fig:full_signal}
\end{subfigure}%
\hfill
\begin{subfigure}{0.48\textwidth}
  \centering
  \resizebox{0.6\textwidth}{!}{
  \includegraphics{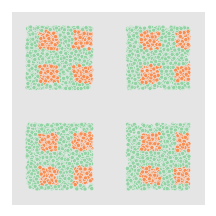}
  }
  \caption{We removed the central line of blocks in both the $x$ and $y$ directions for {Blitzkrige's} training set. 
  }
  \label{fig:pruned_signal}
\end{subfigure}
\\
\begin{subfigure}{\textwidth}
  \centering
  \resizebox{0.5\textwidth}{!}{
  \includegraphics{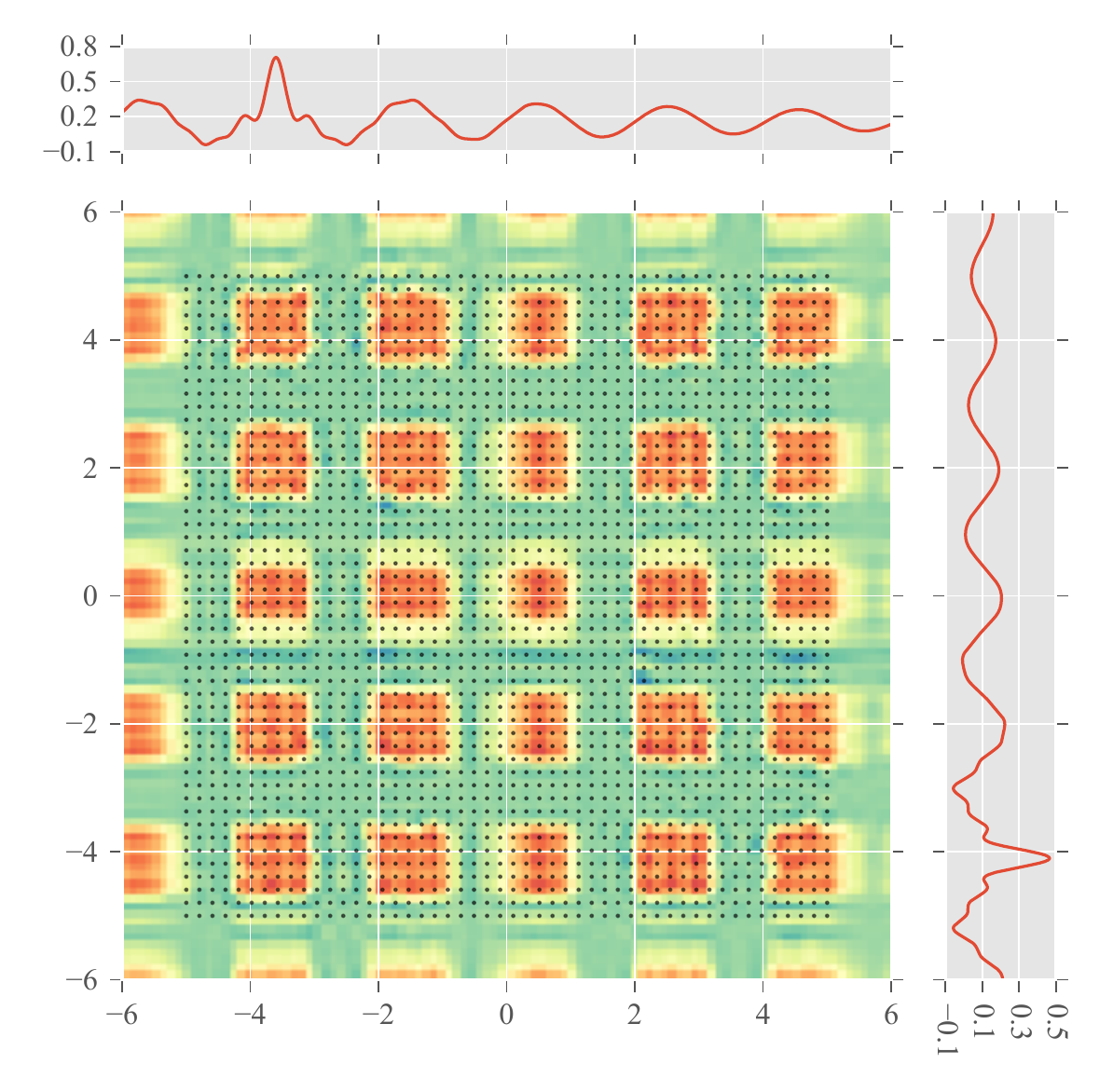}
  }
\caption{Signal prediction and learned kernels: The $50 \times 50$ inducing grid is shown as black dots. The model learns a sum of several local and periodic components, correctly fitting both the sharp discontinuities and the global periodic pattern. The top and right side graphs are the response of the learned \gls{sm} kernel evaluated at the centre of a block in each dimension. }
  \label{fig:prediction}
\end{subfigure}
\caption{Periodic signal reconstruction from non-gridded data. }
  \label{fig:signal}
\end{figure}

Our results for interpolation and extrapolation on a simple two dimensional signal are shown in \cref{fig:signal}. For this signal, we drew 10,000 uniformly distributed points at which we evaluated a 2d square wave function (\cref{fig:full_signal}), and cut out two lines of blocks by excising strips from the middle (\cref{fig:pruned_signal}). We used a $50 \times 50$ grid of inducing points with a 50 element \gls{sm} kernel and learned using Adadelta on minibatches. 
Initially, only inducing points near to data are learned, with the others remaining close to 0. After a few hundred iterations, the model began learning a periodic covariance and hallucinating higher mean values in the regions without data to fit. The results after 1,500 iterations are in \cref{fig:prediction}, showing that the model has filled the signal voids in and extrapolated beyond the inducing point support.


\section{Performance}
\label{perf}

We compare our model against a random forest regressor \cite{breiman2001random} (a successful regression technique for large data sets), the full \gls{gp} and the \gls{svgp} (both from GPy \cite{gpy2014}). For the random forest, we increased the number of estimators until we found no additional benefit, generally between 100 and 1,000. In these tests, we found that additional inducing points or spectral components did not reduce performance once the model was trained, but did impose a runtime cost on training.



\begin{figure}
\centering
\begin{minipage}{0.48\textwidth}
  \resizebox{1.0\textwidth}{!}{
  \includegraphics{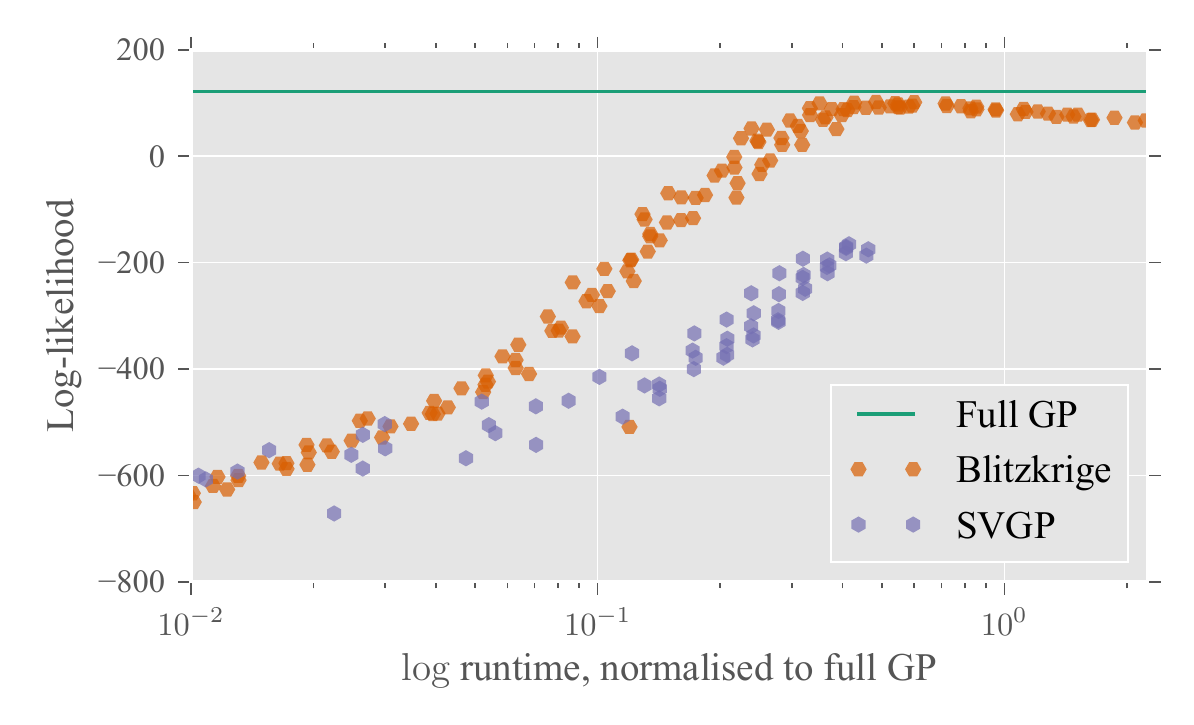}
  }
  \caption{Likelihood compared to run time for {Blitzkriging} and the \gls{svgp} from GPy on samples drawn from a \gls{gp}. 
  }
  \label{fig:perf}
\end{minipage}%
\hfill
\begin{minipage}{0.48\textwidth}
\centering
	\begin{tabular}{lrr} 
\toprule
Model & \acrshort{rmse} &    Log-lik\\ 
\midrule 
\gls{gp} \gls{eq}& \textbf{3.96} & \textbf{0.0371}  \\ 
Blitzkriging \gls{eq} & 4.01 &  -0.0488\\
\bottomrule
\end{tabular}
\captionof{table}{Predictive performance (\gls{rmse} and average predictive log-likelihood) of a full \gls{gp} compared to {Blitzkriging} on four dimensional data from a combined cycle power plant, using the same kernel. }
\label{tab:powerplant}
\end{minipage}
\end{figure}

\subsection{Power plant data}
We compare the performance of our model with that of the \gls{svgp} and a full \gls{gp} on data from a combined cycle power plant, collected and analysed by Heysem, Tüfekci and Gürgenin \cite{combinedcycle2}. This is a 4 dimensional dataset with only 9,568 instances, allowing us to compare the performance in terms of \gls{rmse} and predictive log-likelihood between our model and the full \gls{gp}. With an \gls{eq} kernel and a $7 \times 7 \times 7 \times 7$ inducing grid (a total of 2,401 inducing points) {Blitzkriging} provided a close approximation to the full \gls{gp} using the same kernel. In these tests, there was no change in the performance over multiple restarts. We do not include results for the \gls{svgp} here, because despite multiple restarts and inducing point selection techniques we were unable to find or optimise towards a good inducing point configuration. This is a key benefit of our technique: the dense grids possible with {Blitzkriging} remove the need to perform this difficult and tedious task.


\subsection{Auto-regressive wind prediction}
We trained an \gls{ar}-6 model on wind speed data, on a dataset used by Taylor, McSharry and Buizza \cite{windpower}. We compared with an \gls{arkf} model and our usual random forest. We used a 6 dimensional grid with 3 inducing points per dimension, for a total of 729 inducing points, with either an \gls{eq} or 10 element \gls{sm} kernel at each point. We iteratively split the dataset into six training examples and a one step look ahead prediction, and trained on the first 20,000 of these. 
\Cref{tab:wind} shows the results of our models predicting the remaining 20,174. Blitzkriging handily beats the \gls{arkf} and provides a better mean prediction than a random forest. Again, we do not include results from the \gls{svgp} due to the difficulty of initialising the inducing points. 

\begin{table}
\begin{minipage}[t]{0.48\textwidth}
\centering
\begin{tabular}{lrr} 
\toprule
Model & \Gls{rmse} &   Log-lik\\ 
\midrule 
\Gls{arkf} & 0.699 &  -1.12 \\
Blitzkriging \acrshort{sm} 10 & \textbf{0.645} & -0.990 \\
Blitzkriging \gls{eq} & 0.648  & -0.988  \\
Random Forest & 0.669 & \textbf{-0.921} \\
\bottomrule
\end{tabular}
\caption{Predictive performance on a wind speed autoregressive  task, predicting the next speed using the previous six. }
\label{tab:wind}
\end{minipage}
\hfill
\begin{minipage}[t]{0.48\textwidth}
	\centering
\begin{tabular}{lrr} 
\toprule
Model & \Gls{rmse} &   Log-lik\\ 
\midrule 
Blitzkriging \acrshort{sm} 50 & 0.345 & \textbf{-0.933} \\
\Gls{svgp} sum kernel  & 0.389  & -0.95 \\
Random Forest & \textbf{0.248} & -1.21 \\
\bottomrule
\end{tabular}
\caption{Performance of {Blitzkriging} on two dimensional housing data, compared to the Random Forest and the \gls{svgp}.}
\label{tab:house}
\end{minipage}
\end{table}

\subsection{Housing data}
We replicate the house prices experiment from Hensman et al. \cite{bigdatagp}. We use a $50 \times 70 $ grid of inducing points with a 50 element \gls{sm} kernel.  We use similar data to Hensman et al \cite{bigdatagp}, and initialise and parameterise the \gls{svgp} kernels as in that paper. In our experiment, we use data from the entirety of 2012, and select for semi-detached houses. Selecting for flats (apartments) gives a distribution sharply concentrated in London, while our data-set has more interesting regional structure, testing the ability of the models to learn a combination of local and global structure. In total, we have 172,563 examples and hold back a randomly selected set of 10,000 for test. Before training and for all models, we take the log of the prices, remove the means and normalise. Both \gls{rmse} and predictive log-likelihood are given in this transformed space.

Our model beats the random forest in predictive log-likelihood, however provides a higher mean error. We believe this can be attributed to the non stationarity of the random forest model, which can much more easily fit the sharp changes in house prices around London. Both models beat the \gls{svgp}, showing that our rich \gls{sm} kernel and surfeit of inducing points overwhelms the effect of our simplified variational covariance.
\section{Discussion}
For a machine learning method to be useful, there are several conditions which must be met. It must be scalable, trainable without special knowledge or skills and be capable of discovering structure from data. {Blitzkriging} meets these three desiderata: we scale linearly with data, and allow the use of stochastic gradient descent for large data sets; we rely on off-the-shelf optimisation algorithms (\gls{lbfgs} \cite{Morales:2011:RLL:2049662.2049669} and Adadelta \cite{DBLP:journals/corr/abs-1212-5701}) that reliably find a good parameter fit from a deterministic initialisation, in contrast to the difficulty of fitting the inducing points and kernel parameters in the standard \gls{svgp}; and we do not force the input data to have any specific structure but are still able to exploit the rich \gls{sm} kernel for pattern discovery.

We have shown that Blitzkriging is more robust than the \gls{svgp} and offers better performance at lower computational cost. We have also shown that it is comparable to the random forest on large data sets, improving performance on some metrics.  Some of the random forest's performance can be attributed to its non-stationarity. General inducing point methods can emulate this non-stationarity by changing the density of the inducing points \cite{Titsias09variationallearning}, however, in this paper they are placed on regular grids and not optimised so can not approximate non-stationarity in this way. We may be able to a-priori place grid lines more optimally while avoiding optimisation.

In high dimension and with many inducing points, our requirement to learn a mean value for each point will cause learning to become impractical. For example, learning a function in 100 dimensions with 10 inducing points per dimension would require the specification of $10^{100}$ values. However, the mathematics underlying Blitzkriging  allow a low-rank approximation of the mean to be learned instead, trading off accuracy for further computational gains.

In \cref{fig:signal}, we show that our model can exploit the \gls{sm} kernel for pattern discovery and kernel learning, however we  note that this is currently limited to patterns that can be expressed by independent axis aligned periodicities. This problem could be mitigated by learning an orthonormal rotation matrix $\bs{R} \in \mathbb{R}^{D \times D}$ jointly with the model parameters to project the data into an axis aligned space. This projection, transforming input location $\bs{x}  \rightarrow \bs{R}\bs{x}$, is exactly equivalent to using the Mahalanobis distance  (Vivarelli and Williams \cite{vivarelliwilliams99}) but allows the exploitation of Kronecker structure. Further, we could solve the dimensionality problem mentioned above by following Titsias and L\'{a}zaro-Gredilla \cite{titsias_maha} and learn a projection matrix $\bs{W} \in \mathbb{R}^{D \times K}$ where $K < D$ to project into a lower dimensional subspace.

\small{
\bibliographystyle{unsrt}
\bibliography{bib}
}

\end{document}

%% file: gloss.tex
\newacronym{gp}{GP}{Gaussian process}
\newacronym{svgp}{SVGP}{stochastic variational \gls{gp}}
\newacronym{gpu}{GPU}{graphics processing unit}
\newacronym{gpgpu}{GPGPU}{general purpose graphics processing unit}
\newacronym{ard}{ARD}{automatic relevance detection}
\newacronym{fitc}{FITC}{fully independent training conditional}
\newacronym{se}{SE}{squared exponential}
\newacronym{blas}{BLAS}{basic linear algebra subprograms}
\newacronym{gcd}{GCD}{Grand Central dispatch}
\newacronym{kpca}{KPCA}{kernel principle component analysis}
\newacronym{lbfgs}{L-BFGS}{limited memory Broyden–Fletcher–Goldfarb–Shanno}
\newacronym{rmse}{RMSE}{root mean squared error}
\newacronym{ar}{AR}{auto-regressive}
\newacronym{arkf}{ARKF}{auto-regressive Kalman filter}
\newacronym{eq}{EQ}{exponentiated quadratic}
\newacronym{rf}{RF}{random forest}
\newacronym{kl}{KL}{Kullback–Leibler}
\newacronym{lvm}{LVM}{latent variable model}
\newacronym{gplvm}{GPLVM}{Gaussian process latent variable model}
\newacronym{sm}{SM}{spectral mixture}